\documentclass[pdflatex,sn-mathphys-num]{sn-jnl}

\usepackage{graphicx}%
\usepackage{multirow}%
\usepackage{amsmath,amssymb,amsfonts}%
\usepackage{amsthm}%
\usepackage{mathrsfs}%
\usepackage[title]{appendix}%
\usepackage{xcolor}%
\usepackage{textcomp}%
\usepackage{manyfoot}%
\usepackage{booktabs}%
\usepackage{algorithm}%
\usepackage{algorithmicx}%
\usepackage{algpseudocode}%
\usepackage{listings}%
\usepackage{subcaption}
\usepackage{graphicx} 
\usepackage{rotating} 

\theoremstyle{thmstyleone}%
\theoremstyle{thmstyletwo}%

\theoremstyle{thmstylethree}%

\begin{document}

\title[Adapt, Agree, Aggregate: Semi-Supervised Ensemble Labeling for Graph Convolutional Networks]{Adapt, Agree, Aggregate: Semi-Supervised Ensemble Labeling for Graph Convolutional Networks}


\author*[1]{\fnm{Maryam} \sur{Abdolali}}\email{maryam.abdolali@kntu.ac.ir}

\author[2]{\fnm{Romina} \sur{Zakerian}}\email{romina.zakerian@aut.ac.ir}

\author[2]{\fnm{Behnam} \sur{Roshanfekr}}\email{b.roshanfekr@aut.ac.ir}

\author[2]{\fnm{Fardin} \sur{Ayar}}\email{fardin.ayar@aut.ac.ir}

\author[2]{\fnm{Mohammad} \sur{Rahmati}}\email{rahmati@aut.ac.ir}

\affil[1]{\orgdiv{Department of Mathematics and Computer Science}, \orgname{K. N. Toosi University of Technology}, \orgaddress{\city{Tehran}, \country{Iran}}}

\affil[2]{\orgdiv{Department of Computer Engineering}, \orgname{Amirkabir University of Technology (Tehran Polytechnic)}, \orgaddress{\city{Tehran}, \country{Iran}}}


\abstract{In this paper, we propose a novel framework that combines ensemble learning with augmented graph structures to improve the performance and robustness of semi-supervised node classification in graphs. By creating multiple augmented views of the same graph, our approach harnesses the "wisdom of a diverse crowd", mitigating the challenges posed by noisy graph structures. Leveraging ensemble learning allows us to simultaneously achieve three key goals: adaptive confidence threshold selection based on model agreement, dynamic determination of the number of high-confidence samples for training, and robust extraction of pseudo-labels to mitigate confirmation bias. Our approach uniquely integrates adaptive ensemble consensus to flexibly guide pseudo-label extraction and sample selection, reducing the risks of error accumulation and improving robustness. Furthermore, the use of ensemble-driven consensus for pseudo-labeling captures subtle patterns that individual models often overlook, enabling the model to generalize better. Experiments on several real-world datasets demonstrate the effectiveness of our proposed method.}




\maketitle

\section{Introduction}\label{sec1}
Graphs are ubiquitous data structures for modeling complex relationships between entities in a wide range of real-world domains, such as social networks \cite{li2023survey}, biological systems \cite{bongini2022biognn}, and recommender systems \cite{gao2023survey}.
In recent years, Graph Neural Networks (GNNs) have emerged as a powerful approach for learning directly from graph-structured data \cite{wu2020comprehensive}. By leveraging both node features and the connectivity patterns within the graph, GNNs capture both the underlying structure of the graph and properties of the data simultaneously.
Among the numerous tasks tackled by GNNs, node classification is considered as a core task which aims to assign labels to individual nodes based on both their attributes and their position within the graph. This is particularly important in semi-supervised learning (SSL), where only a few nodes are labeled, and the model must propagate information through the graph to label the majority of unlabeled nodes~\cite{song2022graph}.

The scarcity of labeled data in semi-supervised learning,  makes it difficult for models to learn accurate representations. Additionally, noise in the graph structure, such as incorrect or missing edges, can lead to poor generalization and inaccurate predictions \cite{dai2022towards}. A common solution to these challenges is pseudo-labeling, where the model uses its own predictions to assign labels to the unlabeled nodes \cite{li2023informative,wang2024deep}. By treating high-confidence predictions as pseudo-labels, the model increases the amount of labeled data, helping it learn more effectively.
The ability to use both the graph?s topology and node features makes pseudo-labeling particularly effective for graph-based tasks~\cite{yu2022label}.

High-confidence pseudo-labeling has become a widely adopted approach in SSL, particularly in fields like computer vision \cite{li2022pseco, xiong2021multiview}. The rationale behind this approach is that the more confident a model is about a prediction, the higher the likelihood that it is correct. Therefore, using high-confidence pseudo-labels can significantly enhance the model?s ability to generalize, especially when labeled data is scarce or expensive to obtain.

However, pseudo-labeling does come with its challenges:

\textbf{Error Propagation-} One major concern is error propagation/confirmation bias, where incorrect pseudo-labels assigned in earlier stages of training could negatively impact the model?s performance in later iterations \cite{rizve2021defense}. Since the model relies on its own predictions to label the majority of the data, early mistakes may lead to a cascade of errors as the model refines its parameters based on those incorrect labels. In turn, these errors can reinforce themselves, as the model updates its weights to fit the incorrect pseudo-labels, which can undermine its ability to generalize well to unseen data. This cycle of reinforcing errors is particularly problematic when working with sparse or noisy graphs, where incorrect edges?such as those connecting nodes from different classes?or node features that are inconsistent with those of the same class can introduce significant noise. Over time, these errors may become more pronounced, leading to poorer overall performance.

\textbf{Confidence Threshold Dilemma-} Another critical challenge lies in the selection of high-confidence pseudo-labels \cite{cascante2021curriculum, zhang2021flexmatch}. Determining an appropriate confidence threshold is neither intuitive nor straightforward, as it must balance accuracy with the need for informative samples. A poorly chosen threshold may result in the inclusion of noisy labels, leading to error reinforcement, or in the exclusion of valuable data points, limiting the model's capacity to generalize. Common approaches, such as using fixed thresholds~\cite{wang2021confident} or decreasing the threshold with linear scheduling~\cite{luo2023toward}, fail to account for the specific properties of the input graph structures and cannot adapt to the learning pace.

\textbf{High-Confidence Myopia-} Additionally, relying exclusively on high-confidence pseudo-labels, while seemingly a good strategy, can also be limiting \cite{liu2022confidence}. Although high-confidence labels are typically more accurate and more likely to reflect the true underlying patterns in the data, focusing only on these labels can exclude potentially valuable information. Using only high-confidence labels does not necessarily add much new information to the already available labeled data, especially when these labels are highly similar to the existing ones. This can limit the model's ability to learn from diverse and less certain data points, which could offer unique insights or help the model generalize better. In fact, restricting the model's learning to only the most confident labels may result in overfitting to a narrow subset of the data, preventing the model from exploring the broader structure of the graph and missing out on information that could improve its overall performance. 
Recent work has also highlighted that GNNs can exhibit under-confidence in their predictions. In particular, \cite{wang2021confident} shows that a considerable fraction of correctly classified nodes often receive low confidence scores. This implies that many \emph{true positives} fail to be detected if one relies solely on confidence thresholds. Consequently, the confidence score alone may be a misleading criterion for identifying reliable predictions, underscoring the need for additional methods to assess the trustworthiness of GNN outputs.


To address the limitations of high-confidence pseudo-labeling, we propose A3-GCN, an ensemble-based approach designed to achieve three key objectives. The first 'A' stands for Adaptation, where the ensemble dynamically adjusts the confidence threshold and selects pseudo-labels for each individual GNN based on ensemble output. The second 'A' represents Agreement, leveraging consensus among diverse models as a reliable measure of confidence. The final 'A' stands for Aggregation, where the outputs of individual models are combined to train a robust consensus model, effectively reducing noise, mitigating error propagation, and improving generalization.

The contribution of this work is threefold:
\begin{itemize}
	\item \textbf{Adaptive Confidence Mechanism:} We introduce a novel approach for selecting confidence thresholds dynamically based on ensemble model agreement, overcoming limitations of static thresholding.
	
	\item \textbf{Dynamic Sample Selection:} Our framework adaptively determines the number of high-confidence samples for training, enhancing model robustness and generalization by incorporating diverse informative data points.
	
	\item \textbf{Ensemble-Driven Robustness:} Through a consensus-based pseudo-labeling strategy, we capture subtle patterns overlooked by individual models.
\end{itemize}

The paper is organized as follows. In Section 2, we provide the necessary preliminaries and formally define the problem, laying the foundation for the proposed approach. Section 3 offers a comprehensive review and discussion of related works in the area of SSL for graphs, with a focus on recent advancements and challenges. In Section 4, we present our novel ensemble-based approach for leveraging high-confidence pseudo-labels in graph neural networks, highlighting the key components and mechanisms. We investigate the properties and performance of proposed approach using synthetic and real-world data sets in Section 5.

\section{Preliminaries and Problem Formulation}\label{sec2}

This section covers the fundamentals of graph neural networks, introduces essential terminology, and explains core graph-related concepts. 

\subsection{Graph Fundamentals}
A graph \( G = (V, E) \) consists of a set of nodes \( V = \{v_1, v_2, \dots, v_N\} \) and edges \( E \subseteq V \times V \), where \( N = |V| \) denotes the number of nodes. The graph structure is often represented by an adjacency matrix \( A \in \mathbb{R}^{N \times N} \), where \( A_{ij} = 1 \) if there is an edge between nodes \( v_i \) and \( v_j \), and \( A_{ij} = 0 \) otherwise. For weighted graphs, \( A_{ij} \) indicates the weight of the edge. 

Each node \( v_i \) is associated with a feature vector \( x_i \in \mathbb{R}^d \), where \( d \) is the dimensionality of the feature space. Collectively, these features are represented by a node feature matrix \( X \in \mathbb{R}^{N \times d} \). Additionally, nodes belong to one of \( C \) classes, and their labels are denoted by \( Y \in \mathbb{R}^{N \times C} \), where \( Y_i \) is the one-hot encoded label vector for node \( v_i \).

\subsection{Graph Neural Networks}
Graph Neural Networks (GNNs) leverage both graph structure and node features to perform tasks such as node classification, link prediction\cite{zhang2018link}, and graph classification\cite{defferrard2016convolutional}. A fundamental GNN model, the Graph Convolutional Network (GCN), performs message passing by aggregating information from neighboring nodes\cite{kipf2016semi}. The forward pass of a GCN layer is defined as:
\begin{equation}
	H^{(l+1)} = \sigma\left(\tilde{D}^{-1/2} \tilde{A} \tilde{D}^{-1/2} H^{(l)} W^{(l)}\right),
\end{equation}

In this expression, \(H^{(l)} \in \mathbb{R}^{N \times d_l}\) represents the node embedding matrix at layer \(l\) with the initial embeddings given by \(H^{(0)} = X\). The trainable weight matrix \(W^{(l)} \in \mathbb{R}^{d_l \times d_{l+1}}\) linearly transforms these embeddings to produce features in a new space of dimension \(d_{l+1}\). The matrix \(\tilde{A} = A + I\) is the adjacency matrix with self-loops added, ensuring that each node is connected to itself, while \(\tilde{D}\) is the diagonal degree matrix of \(\tilde{A}\) with entries \(\tilde{D}_{ii} = \sum_j \tilde{A}_{ij}\). The symmetric normalization given by \(\tilde{D}^{-1/2} \tilde{A} \tilde{D}^{-1/2}\) ensures that the features are properly scaled when aggregated from neighboring nodes. Finally, the non-linear activation function \(\sigma(\cdot)\) is applied element-wise to introduce non-linearity into the model.

In semi-supervised learning (SSL) for node classification, the goal is to predict the labels of unlabeled nodes using both labeled and unlabeled data. Given a graph \( G \), the dataset is divided into labeled nodes \( V_L \subset V \) with corresponding labels \( Y_L \), and unlabeled nodes \( V_U = V \setminus V_L \). The task is to learn a function \( f: \mathbb{R}^{N \times d} \times \mathbb{R}^{N \times N} \to \mathbb{R}^{N \times C} \) that predicts the labels \( Y_U \) of the unlabeled nodes.

\section{Related work}\label{sec3}
In this section, we provide an overview of related works in the domain of semi-supervised learning, with a focus on graph-based semi-supervised learning approaches. We also discuss self-training methods, which have been widely employed to improve the performance of GNNs in scenarios with limited labeled data.
\subsection{Semi-Supervised learning}
Semi-Supervised Learning that has recently gained significant attention, is a machine learning approach that bridges the gap between supervised and unsupervised learning by utilizing both labeled and unlabeled data \cite{CascanteBonilla2020CurriculumLR}. SSL aims to improve model performance by effectively exploiting the vast amount of unlabeled data alongside a limited set of labeled examples.
Key methodologies in SSL include: 1) Self-Training: The model iteratively trains on its own predictions for the unlabeled data, treating them as pseudo-labels, which are refined as the training progresses \cite{CascanteBonilla2020CurriculumLR, Lee2013PseudoLabelT}. For example Chen Gong et al. \cite{7465792} leverages curriculum learning to classify unlabeled images by evaluating their difficulty based on reliability and discriminability. Using a Multi-Modal Curriculum Learning (MMCL) strategy, multiple feature-specific "teachers" analyze image difficulty, reach a consensus, and guide the learner to classify images in a sequence from simple to difficult.
2) Consistency Regularization: This method enforces the model to produce consistent predictions under small perturbations to the input data, leveraging the assumption that similar inputs should yield similar outputs \cite{10.5555/3454287.3454741, NEURIPS2020_06964dce}.

\subsection{Graph-based Semi-supervised Learning}

Graph-based semi-supervised learning is a powerful approach that leverages the structure of data represented as a graph to propagate label information from a small set of labeled nodes to a large set of unlabeled nodes \cite{yu2022label}. The key idea is to exploit the graph's topology to infer labels for unlabeled nodes, under the assumption that connected nodes are likely to share similar labels. Methods like Label Propagation (LP) and Graph Neural Networks are commonly used for this purpose. These approaches often incorporate additional techniques such as regularization, curriculum learning, or multi-modal integration to improve performance, especially in cases with limited labeled data. For example Yu et al. \cite{8501586} examines pairwise relationships in the latent space and leverages the graph structure to facilitate semantic learning in a semi-supervised context. While they employ a traditional feature selection framework, their method utilize GNNs for semi-supervised node classification. About using curriculum learning technique, Chen Gong et al.\cite{7447818} proposed a novel approach to label propagation by treating unlabeled examples with varying levels of difficulty. The method assesses the reliability and discriminability of examples and optimizes the propagation process by progressing from simpler to more complex cases. Another method that is using curriculum learning to guide Label Propagation on graphs has done with his team again \cite{10.1145/3322122}. For multi-modal data, modality-specific "teachers" evaluate example difficulty and collaborate to select the simplest examples for propagation. Common preferences among teachers are captured in a row-sparse matrix, while their distinct views are modeled with a sparse noise matrix, enabling efficient and adaptive label learning. Traditional approaches to semi-supervised often rely on graph Laplacian regularization techniques \cite{JMLR:v7:belkin06a, 10.1145/1390334.1390438,10.5555/2981345.2981386}. For instance, Belkin et al. \cite{JMLR:v7:belkin06a} utilize the geometry of the marginal distribution to introduce a novel form of regularization. Existing GNN methods primarily concentrate on designing efficient message-passing mechanisms \cite{kipf2016semi,wu2019simplifying} but often fail to fully utilize the information from unlabeled nodes. In contrast, our work addresses this limitation by employing effective pseudo-labeling to enhance semi-supervised node classification.

\subsection{Self-training}
Graph Neural Networks (GNNs) have achieved success in graph modeling, but they often face the challenge of over-smoothing, where representations of graph nodes from different classes become indistinguishable as more layers are added. This issue negatively impacts model performance, such as classification accuracy \cite{DBLP:conf/aaai/ChenLLLZS20}. To address this, Deli Chen et al. propose a method called Adaptive Edge Optimization (AdaEdge2), which alleviates over-smoothing by dynamically adjusting the graph's structure \cite{DBLP:conf/aaai/ChenLLLZS20}. Through an iterative process, the method trains GNN models while removing or adding edges based on predictions, tailoring the graph for the specific learning task. Experimental results demonstrate that AdaEdge2 effectively reduces over-smoothing and enhances overall model performance.

Qimai Li et al. also released a paper that addresses the over-smoothing problem in GCNs and examines its rapid onset in small datasets, where even a few convolutional layers can cause feature mixing \cite{10.5555/3504035.3504468}. Moreover, adding layers increases the difficulty of training. While shallow GCNs, such as the two-layer model, avoid over-smoothing, they come with their own limitations.
To address these challenges and fully realize the potential of GCNs, the authors propose two approaches: The co-training approach pairs a GCN with a random walk model, leveraging the latter to enhance the exploration of global graph topology. Meanwhile, the self-training approach harnesses the GCN's feature extraction capabilities to address its limitations stemming from localized filters.

Ke Sun et al. address the challenge of improving label propagation in GCNs with limited labeled data \cite{sun2020multi}. To achieve better generalization, deeper GCN architectures are suggested for spreading weak label signals more effectively. However, to overcome the inefficiency of shallow GCNs in propagating label information, the authors propose a framework integrates DeepCluster \cite{caron2018deep}, a self-supervised learning method, into the graph embedding process. A novel cluster-aligning mechanism is introduced to generate pseudo-labels for unlabeled data in the embedding space, aiding in classification tasks. By seamlessly combining DeepCluster and the aligning mechanism with the Multi-Stage Training Framework, the authors formally propose the Multi-Stage Self-Supervised (M3S) Training Algorithm.

From another aspect, several studies have demonstrated that pseudo-labeling can enhance the performance of graph learning models. However, incorrect pseudo-labels may significantly harm the training process, particularly on graph data, where errors can propagate through the network. To address this gap, Wang et al. \cite{botao2023deep} provide a detailed theoretical analysis of pseudo-labeling in graph learning. The authors show that the error in pseudo-labeling is influenced by the confidence threshold and the consistency of multi-view predictions and examine how pseudo-labeling impacts the convergence properties of graph learning models. Based on their findings, a cautious pseudo-labeling strategy is proposed, which focuses on labeling only those samples with the highest confidence and consistency across multiple views~\cite{botao2023deep}.

Hu et al. \cite{10.1145/3459637.3482469} investigate two challenges: Labeling non-Euclidean data is typically more expensive than labeling Euclidean data, yet many existing graph convolutional networks (GCNs) rely on limited labeled data while neglecting the potential of unlabeled data. To address these challenges, they propose a novel end-to-end framework called Iterative Feature Clustering Graph Convolutional Networks (IFC-GCN) that incorporates an Iterative Feature Clustering (IFC) module, which refines node features iteratively using predicted pseudo-labels and feature clustering. Additionally, an EM-like training framework is introduced to alternately refine pseudo-labels and node features, effectively boosting the network?s performance.

Han Yang et al. proposed Propagation-regularization (P-reg), a novel method that enhances GNN performance by providing additional supervisory signals to nodes \cite{yang2021rethinking}. P-reg is as powerful as an infinite-depth GCN, enabling long-range information propagation without over-smoothing or high computational costs and make the model predictions of each node?s neighbors to supervise itself.

Graph convolutional networks (GCNs) often struggle to utilize unlabeled data effectively, especially for distant nodes. To address this, Luo et al. \cite{luo2020every} proposes the Self-Ensembling GCN (SEGCN) that integrates GCN with the Mean Teacher framework. SEGCN uses a student-teacher model, where the student learns from labeled data and aligns with the teacher's predictions on unlabeled data, even under challenging conditions. The teacher, in turn, averages the student?s weights to provide more accurate guidance. This iterative process effectively leverages both labeled and unlabeled data, enhancing GCN training.

\section{Proposed method: A3-GCN}\label{sec4}
\begin{figure}[h]
	\centering
	\includegraphics[width=1\textwidth]{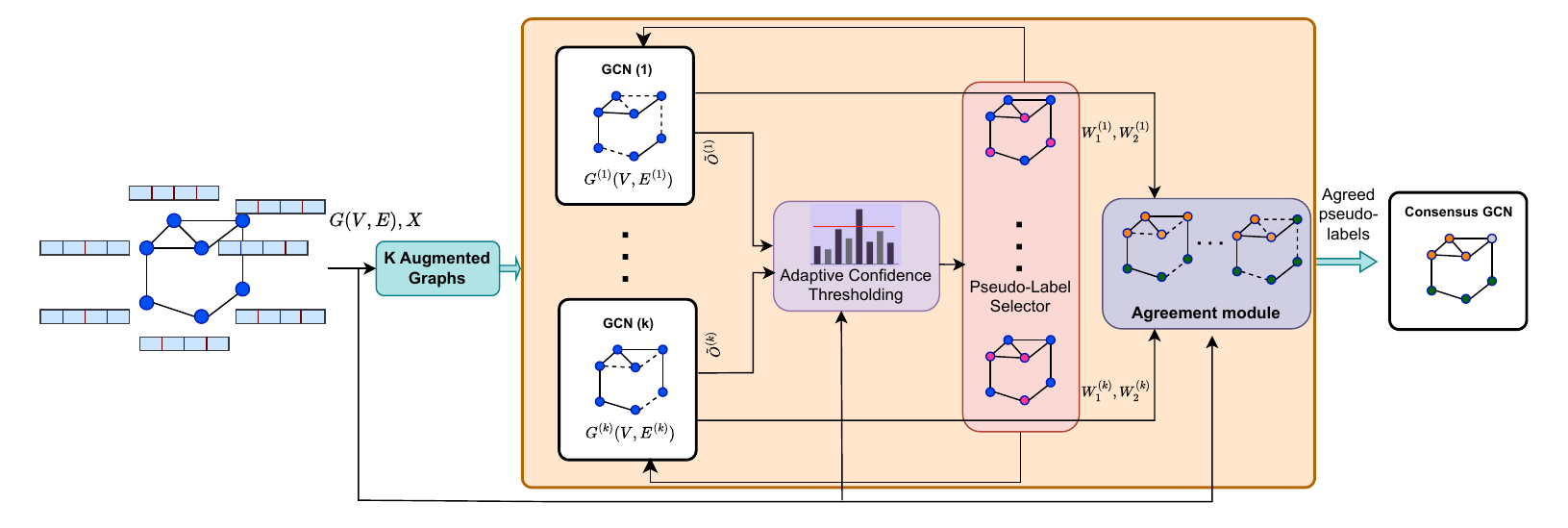}
	\caption{Overall schematic of our approach. The given graph $G(V, E)$ is first augmented to generate $k$ similar, yet diverse sets of `views` of the graphs, which are then assigned to $k$ GCN models independently. The feature matrix $X$ remains consistent across all models. In each epoch, the outputs of these models serve two purposes: setting the adaptive threshold for high-confidence samples and selecting a set of nodes to be assigned pseudo-labels. These selected pseudo-labels are used as supervision for each of $k$ models. The outputs are then passed to an agreement module, which identifies the consensus pseudo-labels, which are subsequently provided to a final consensus GCN model along with the original graph.}
	\label{fig:overall}
\end{figure}

We present A3-GCN, a novel SSL framework designed around three key principles: Adaptation, Agreement, and Aggregation.
A3-GCN integrates four key components: (i) generating diverse multi-model GCNs through graph augmentation to address noisy graph structures and leverage ensemble learning, (ii) incorporating adaptive high-confidence pseudo-labels to enhance the supervised signal, (iii) dynamically selecting high-confidence pseudo-labels to mitigate confirmation bias, and (iv) training a final consensus GCN to prevent over-reliance on easy-to-predict pseudo-labels. These components work synergistically to produce a robust model. An illustration of the proposed approach is shown in Figure \ref{fig:overall}. Below, we describe each key components of the method in details.

\subsection{Multi-Model GCNs with Augmentation and Pseudo-Labels}

GCNs rely heavily on the structure of the input graph to propagate and aggregate information across nodes. However, the presence of noisy edges in real-world graphs can significantly degrade the performance of GCNs~\cite{xu2022ned,luo2021learning}. To address this challenge, we use edge drop augmentation to generate diverse graph views and mitigate the impact of noisy edges. By randomly removing a subset of edges in the graph, we effectively reduce the probability that a noisy edge is included in any single augmented graph and generate multiple augmented graphs with slightly different structures. Note that the use of dropout in the GCN model and varying random initializations contribute to the generation of diverse graph structures as well.

For a given graph \( G = (V, E) \), we define an edge drop operation with probability \( p_{\text{drop}} \). For each augmented graph \( G^{(i)} = (V, E^{(i)}) \) ($i=1,...k$), edges are independently sampled from \( E \) such that:
\begin{equation}
	E^{(i)} = \{(u, v) \mid z_{uv} > p_{\text{drop}}, \, z_{uv} \sim \text{Uniform}(0, 1), \,  (u,v) \in E\}.
\end{equation}

The augmented graphs are passed through independent GCNs, each using a shared node feature set \( X \). Each \( \mathrm{GCN} (i) \) has two hidden layers, parameterized by distinct weight matrices \( W_1^{(i)} \) and \( W_2^{(i)} \), where \( i = 1, \dots, k \), corresponding to the \( k \) augmented graph views.
Considering \( \hat{Y}^{(i)}\in \mathbb{R}^{N\times C} \) as the output of the $i$-th GCN model for the input feature set \(X\), we define the confidence of the prediction for a node \(u\in V\) as $\text{Conf}_u^{(i)} = \max\Big(\hat{Y}_u^{(i)}\Big)$, where $\hat{Y}_u^{(i)}$ denotes the probability distribution over classes for node $u$, i.e., the row of \(\hat{Y}^{(i)}\) corresponding to node $u$.
During the training of each individual GCN, the set of labeled data is expanded by incorporating the currently estimated high-confidence pseudo-labels. A node is selected for pseudo-labeling if its confidence exceeds a predefined threshold \( \theta_{\text{conf}}\). More specifically, the set of candidate high-confidence pseudo-labeled nodes for the $i$-th GCN is defined as:
\begin{equation}
	\mathcal{H}_{\text{pseudo}}^{(i)} = \{u \mid \text{Conf}_u^{(i)} \geq \theta_{\text{conf}}, u \in V\}.
\end{equation}
Once high-confidence nodes are identified, their estimated labels could be used to augment the supervised data for the next epoch of the training iteration.

\subsection{Power of Agreement}
High-confidence pseudo-labels can unintentionally introduce confirmation bias into the training process such that pseudo-labels create a feedback loop where the model becomes increasingly confident in its wrong predictions. 
To mitigate confirmation bias and prevent the degradation of performance due to incorrect pseudo-labels, we propose training each individual model on a random subset of high-confidence nodes, where the size of this subset is adaptively determined based on the level of agreement among the models on these high-confidence predictions.  Specifically, when the models exhibit strong agreement on the high-confidence nodes?indicating a consensus that the pseudo-labels are likely reliable?we increase the size of the random subset of high-confidence pseudo-labels for training. Conversely, when the models diverge significantly in their predictions, it suggests that the high-confidence nodes are inconsistent and may not be truly reliable. In such cases, we reduce the size of the random subset, thereby limiting the influence of potentially erroneous pseudo-labels and mitigating confirmation bias.

Let \( \mathcal{H}_{\text{pseudo}}^{(i)} \) be the set of high-confidence nodes for model \( i \). The agreement among models on a given node is quantified using the intersection and union of the high-confidence sets across all models.

The adaptive size \( S_{\text{high-conf}} \) of the high-confidence node subset for training each model is defined as the ratio of the intersection to the union of high-confidence nodes across all models:
\begin{equation} \label{high-ratio}
	S_{\text{high-conf}} = \frac{|\mathcal{H}_{\text{intersect}}|}{|\mathcal{H}_{\text{union}}|},
\end{equation}
where \( \mathcal{H}_{\text{intersect}} = \bigcap_{i=1}^k \mathcal{H}_{\text{pseudo}}^{(i)} \) is the set of nodes that are high-confidence across all models, and \( \mathcal{H}_{\text{union}} = \bigcup_{i=1}^k \mathcal{H}_{\text{pseudo}}^{(i)} \) is the set of nodes that are high-confidence in at least one model.

The adaptive size \( S_{\text{high-conf}} \)  quantifies the degree of agreement among the models. When \( S_{\text{high-conf}} \) is large (close to 1), it indicates that most models agree on the high-confidence nodes, and therefore the random subset of high-confidence nodes for training each model is increased. On the other hand, when \( S_{\text{high-conf}} \) is small (close to 0), indicating significant divergence in the models' predictions, the random subset size is decreased, reducing the risk of reinforcing incorrect pseudo-labels.
This adaptive mechanism ensures that the inclusion of pseudo-labels is guided by the collective reliability of the models, striking a balance between exploiting reliable information and avoiding the amplification of errors through feedback loops. 

\subsection{Dynamic Confidence Thresholding}
The performance of SSL based on pseudo-labels heavily depends on the value of \(\theta_{conf}\). High values of this parameter enforce a strict selection of pseudo-labels, leading to a set that closely mirrors the original training data. On the other hand, low values introduce many incorrect pseudo-labels, which can significantly degrade performance. 

The proposed ensemble approach adaptively adjusts this parameter. Specifically, if the agreement between models continues to increase across successive iterations, it indicates that the models are aligning and agreeing on a larger number of pseudo-labels, thus allowing the threshold to be decreased. Conversely, if the agreement decreases, it signals that the models are diverging and that the pseudo-labels are likely unreliable, prompting an increase in the threshold. We propose updating \(\theta_{conf}\) as follows:

\begin{equation} \label{adaptive_conf}
	\theta_{conf}^{k} \leftarrow \theta_{conf}^{k-1} + \alpha \left(S_{high-conf}^{k-1} - S_{high-conf}^{k}\right),
\end{equation}

where \(\theta_{conf}^{k}\) denotes the confidence threshold and \(S_{high-conf}^{k}\) represents the agreement ratio at the \(k\)-th epoch, respectively and $\alpha$ indicates the learning rate parameter.

\subsection{Consensus Model Training}

While we train the individual models with the adaptive selection of high-confidence pseudo-labels, we combine their predictions to form a more robust final model, which we refer to as the \textit{consensus model}. The consensus model is trained on a combination of nodes for which the majority of the models agree on their labels, including both high-confidence and low-confidence nodes.

Formally, let \( \mathcal{V}_{\text{consensus}} \) represent the set of nodes used to train the consensus model:
\begin{equation}
	\mathcal{V}_{\text{consensus}} = \{ u \mid \text{Agree}(u) \geq \beta \cdot k \}.
\end{equation}
Here, \(\beta \in [0,1]\) is a hyperparameter controlling the majority threshold for agreement, ensuring that the consensus model is trained on nodes for which most of the models agree. This guarantees that the consensus model learns from nodes with consistent and reliable pseudo-labels, thus avoiding the reinforcement of noisy or incorrect labels from divergent models. We set $\beta=1$ for all experiments in this paper.

For each node \( u \in \mathcal{V}_{\text{consensus}} \), the label \( \hat{y}_u^{\text{consensus}} \) is determined by the majority vote among the \( k \) models:
\begin{equation}
	\hat{y}_u^{\text{consensus}} = \text{MajorityVote} \left( \hat{y}_u^{(1)}, \hat{y}_u^{(2)}, \dots, \hat{y}_u^{(k)} \right),
\end{equation}
where \( \hat{y}_u^{(i)} \) represents the predicted label for node \( u \) by model \( i \). 

The consensus model is then trained on the nodes in \( \mathcal{V}_{\text{consensus}} \) using the majority vote labels as the ground truth, with cross-entropy loss:
\begin{equation}
	\mathcal{L}_{\text{consensus}} = - \sum_{u \in \mathcal{V}_{\text{consensus}}} \sum_{j=1}^{C} \hat{Y}_{u}^{\text{consensus}}[j] \log \left( \hat{o}_{u}^{\text{consensus}}[j] \right),
\end{equation}
where \( \hat{o}_{u}^{\text{consensus}} \) represents the softmaxed output of the consensus model, and \( \hat{Y}_{u}^{\text{consensus}}[j] \) is the \( j \)-th probability in the pseudo-label vector for the node \( u \).

This approach ensures that the consensus model benefits from the collective wisdom of the individual models, while minimizing the impact of noisy or inconsistent pseudo-labels by focusing on nodes where agreement is strong.

In addition to the high-confidence nodes, the inclusion of low-confidence nodes with majority agreement helps the consensus model adapt to regions of the graph where the predictions are uncertain but consistent across the models. 
By incorporating agreement as a criterion for selecting pseudo-labels, we move beyond simple confidence thresholds and emphasize predictions that are robust across multiple perspectives. Nodes with high agreement scores are more likely to represent genuinely correct labels, even in the presence of graph noise or ambiguous features. This improves the overall quality of the pseudo-labeled set, ensuring that the training process is guided by reliable supervision.
The algorithm is summarize in Algorithm~\ref{alg:A3-GCN}.

GNNs tend to be under-confident in their predictions, where the prediction accuracy exceeds the model?s confidence~\cite{wang2021confident}. In other words, many correct predictions fall within the low-confidence range, which contrasts with the behavior of most modern deep neural networks \cite{guo2017calibration}. This implies that relying solely on high-confidence thresholding results in the loss of information from low-confidence samples. On the other hand, ensemble learning has been shown to be effective for selecting pseudo-labels \cite{lakshminarayanan2017simple}. To the best of our knowledge, our work is the first to use ensemble learning as a criterion for selecting pseudo-labels in GCNs.

\begin{algorithm}[!ht]
	\caption{A3-GCN: Semi-Supervised Node Classification}
	\label{alg:A3-GCN}
	\begin{algorithmic}[1]
		\State \textbf{Input:} Graph \( G = (V, E) \), node features \( X \), labeled nodes \( \mathcal{L} \), 
		\State \textbf{Initialize:} number of GCN models \( k \), learning rate \( \alpha \),  majority threshold \( \beta \), number of epochs max\_epochs
		\State Generate augmented graphs \( G^{(i)} = (V, E^{(i)}) \) for \(i=1,...,k\)
		\State Initialize \( k \) GCN models for augmented graphs and a consensus model for the original graph
		
		\For{j = 1 \textbf{to} max\_epochs}
		\State Calculate the adaptive subset ratio:
		\( S_{\text{high-conf}} = \frac{|\mathcal{H}_{\text{intersect}}|}{|\mathcal{H}_{\text{union}}|} \)
		
		\For{each model \( i = 1 \dots k \)}
		\State \( \mathcal{L}_{\text{pseudo}}^{(i)} = \{ u \mid \text{Conf}_u^{(i)} \geq \theta_{\text{conf}}, u \in V \} \)
		\State \( \mathcal{L}^{(i)} \leftarrow \mathcal{L} \cup \mathcal{L}_{\text{pseudo}}^{(i)} \)
		\State Train model \( i \) using the augmented graph \( G^{(i)} \), node features \( X \), and a random subset of \( \mathcal{L}^{(i)} \) with size \( S_{\text{high-conf}} \).
		\EndFor
		
		\State Update confidence threshold \( \theta_{\text{conf}} \) based on agreement:
		\[
		\theta_{\text{conf}} \leftarrow \theta_{\text{conf}} + \alpha \left(S_{\text{high-conf}}^{j-1} - S_{\text{high-conf}}^{j}\right)
		\]
		
		\State Identify nodes \( \mathcal{V}_{\text{consensus}} \) where most models agree:
		\[ \mathcal{V}_{\text{consensus}} = \{ u \mid \text{Agree}(u) \geq \beta \cdot k \} \]
		
		\State Set the label \( \hat{y}_u^{\text{consensus}} = \text{MajorityVote} \left( \hat{y}_u^{(1)}, \hat{y}_u^{(2)}, \dots, \hat{y}_u^{(k)} \right) \) for \( u \in \mathcal{V}_{\text{consensus}} \)
		
		\State Train consensus model on \( \mathcal{V}_{\text{consensus}} \) using the majority vote labels as the ground truth
		\EndFor
		\State \textbf{Output:} Trained Consensus GCN model
	\end{algorithmic}
\end{algorithm}

\section{Experimental results}\label{sec5}
In this section, we evaluate and analyze the performance of proposed model. The code is available from \url{https://github.com/mabdolali/selfsupervised_GNN}. 

\paragraph{Datasets:}
The datasets used for evaluation include three citation networks representing real-world data. In the citation networks, nodes represent documents, edges correspond to citation links, and node features are word vectors extracted from the documents. The datasets differ in size and complexity, and their details are summarized in Table~\ref{tab:dataset_summary}.

\begin{table}[h]
	\centering
	\caption{Summary of datasets used for evaluation.}
		\begin{tabular}{|l|c|c|c|c|}
			\hline
			\textbf{Dataset} & \textbf{\#Nodes} & \textbf{\#Edges} & \textbf{\#Features/Node} & \textbf{\#Classes} \\ \hline
			Cora \cite{mccallum2000automating}           & 2,708            & 5,429            & 1,433                   & 7                  \\ \hline
			CiteSeer  \cite{10.1145/276675.276685}      & 3,327            & 4,732            & 3,703                   & 6                  \\ \hline
			PubMed \cite{Sen_Namata_Bilgic_Getoor_Galligher_Eliassi-Rad_2008}         & 19,717           & 44,338           & 500                     & 3                  \\ \hline
		\end{tabular}
	\label{tab:dataset_summary}
\end{table}

\subsection{Beyond easily identifiable high-confidence nodes}
High-confidence nodes are often the primary candidates for reliable predictions. In this experiment, we aim to explore the behavior of high-confidence nodes over multiple training epochs and investigate how nodes that are in agreement across models (i.e., agreed nodes) behave over time.

\begin{figure}[!ht]
	\centering
	\begin{subfigure}[b]{0.45\linewidth}
		\centering
		\includegraphics[width=5cm]{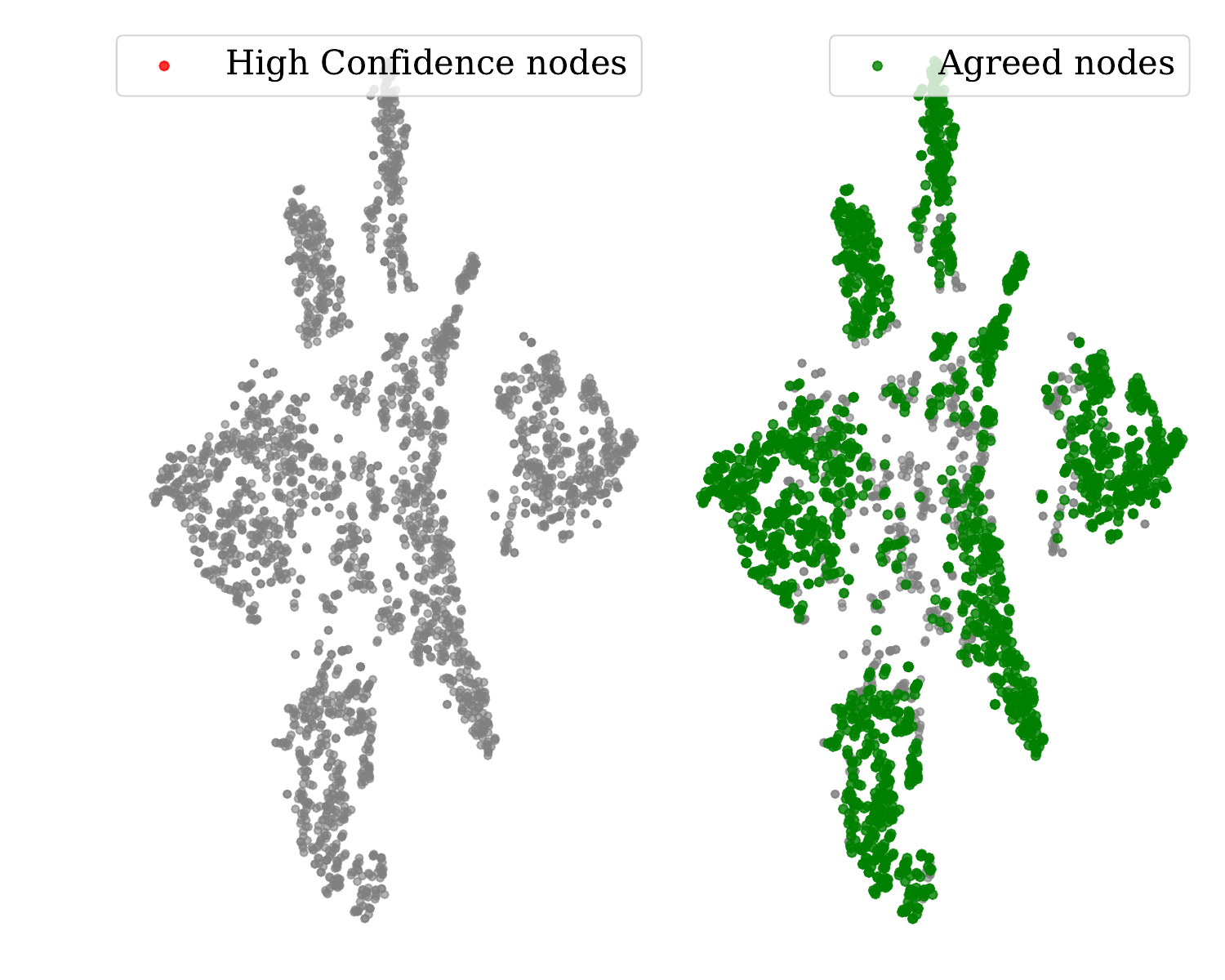}
		\caption{5th Epoch, 88\% of agreed nodes are predicted correctly}
	\end{subfigure}
	\hfill
	\begin{subfigure}[b]{0.45\linewidth}
		\centering
		\includegraphics[width=5cm]{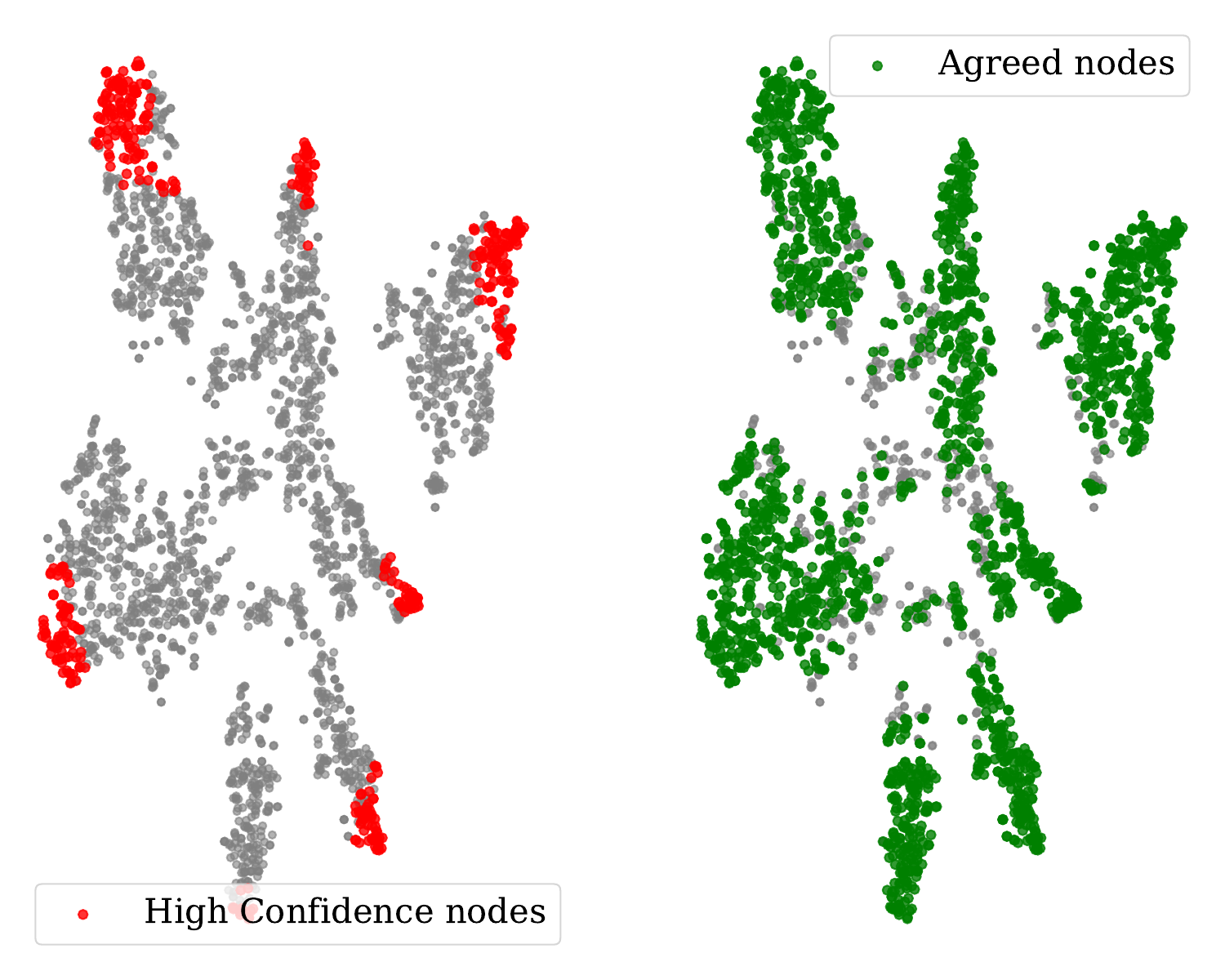}
		\caption{10th Epoch, 89\% of agreed nodes are predicted correctly}
	\end{subfigure}
	\vfill
	\begin{subfigure}[b]{0.6\linewidth}
		\centering
		\includegraphics[width=5cm]{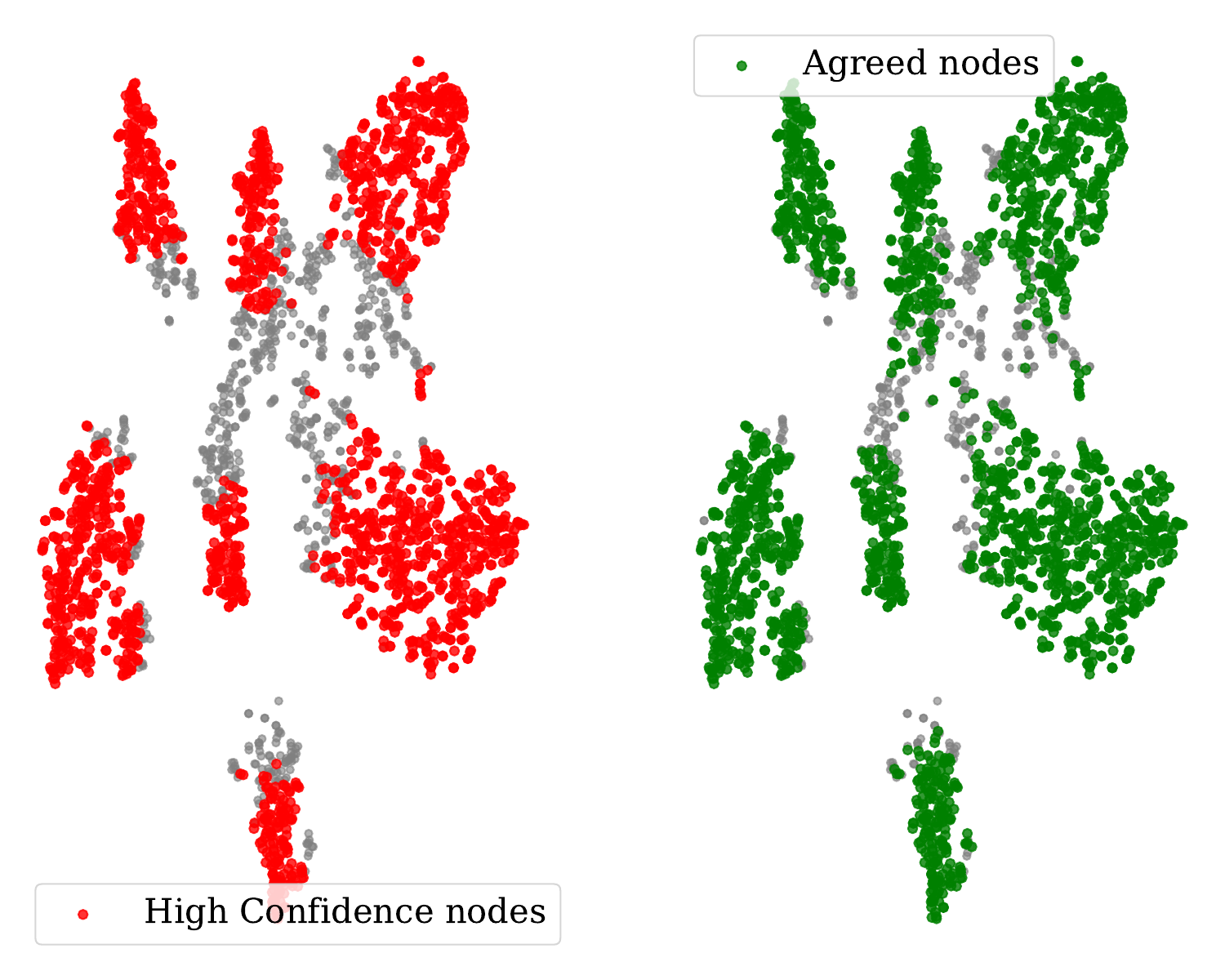}
		\caption{100th Epoch, 91\% of agreed nodes are predicted correctly}
	\end{subfigure}
	\caption{Two-dimensional embeddings of the nodes and the spread of high confident nodes (in red) and agreed nodes (in green).}
	\label{graph}
\end{figure}

We perform this experiment using 2D t-SNE embeddings of nodes at several epochs during the training process. This allows us to visually inspect how the nodes' embeddings evolve and how high-confidence and agreed nodes are distributed in the embedding space. The embeddings at epochs 5, 10, and 100 are plotted in Figure~\ref{graph} as subfigures a, b, and c, respectively. We observe that:
\begin{itemize}
	\item Agreed nodes are more spread out across the embedding space compared to high-confidence nodes, particularly in the earlier epochs. Conversely, high-confidence nodes initially behave conservatively and provide limited informativeness. We observe a distribution shift between the original dataset and the self-trained augmented dataset, a challenge also noted in \cite{liu2022confidence}. It was also observed that relying solely on high-confidence nodes does not always yield high information gain.
	\item As training progresses, the agreed nodes transition into high-confidence nodes, which better guide the supervision. This is expected, as these pseudo-labels serve as supervisory signals.
	
	\item Interestingly, we conducted a similar experiment using only high-confidence pseudo-labels agreed upon by all k models to train the consensus model. This setup contrasts conservative agreement with noisy agreement. Notably, our findings show that the noisy agreement approach (current method) achieves a higher average accuracy of 85.36\% across 10 trials, compared to 84.01\% for the conservative agreement.
	
	\item The proportion of correctly predicted agreed nodes remains consistently high across all epochs. Notably, it starts at 77\% and rapidly increases to around 90\% within the first 10 epochs. This confirms that agreed pseudo-labels are reliably predicted throughout training, providing largely accurate supervision.
\end{itemize}

\subsection{Conservative pseudo-labeling}
\begin{figure}[h]
	\begin{minipage}[b]{1\linewidth}
		\centering
		\centerline{\includegraphics[width=9cm]{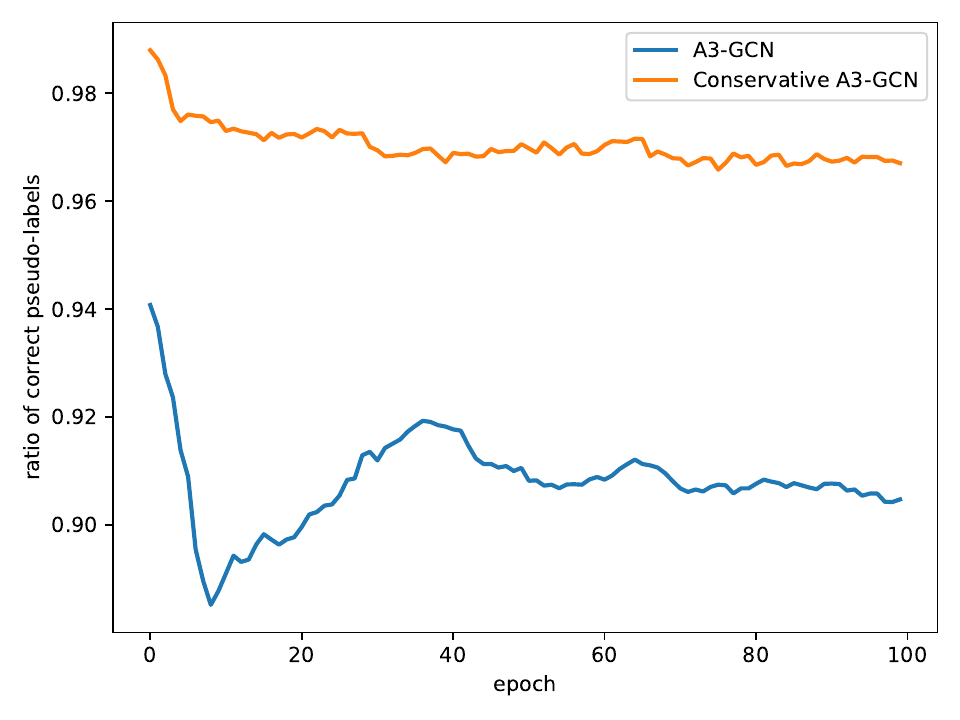}}
	\end{minipage}	%
	\caption{ratio of correct pseudo-labels for A3-GCN and the conservative baseline across epochs.}
	\label{conservative}
\end{figure}

We compared our proposed A3-GCN with a conservative baseline, where we retain all the adaptive configurations of A3-GCN but only use the high-confidence agreed nodes from $k$ models, rather than utilizing all the agreed nodes, to train the final consensus model. The ratio of correct pseudo-labels at each epoch is shown in Figure~\ref{conservative}, where the conservative approach achieves an impressive 98\% correct pseudo-label ratio, compared to 90\% for the original A3-GCN. However, in terms of overall accuracy, the original A3-GCN achieved an average accuracy of 85.40\%, whereas the conservative A3-GCN reached 84.20\% across 10 trials.

The discrepancy in the ratio of correct pseudo-labels and overall accuracy between A3-GCN and the conservative baseline suggests a potential distribution shift in the model?s learning process~\cite{wang2021confident}. Specifically, the conservative approach, which only uses high-confidence agreed nodes, appears to rely on a narrower subset of the data, leading to a higher pseudo-label accuracy (98\%) but lower overall performance (84.20\% accuracy). This indicates that the conservative method might be overfitting to a particular, possibly less diverse distribution of high-confidence nodes. In contrast, the original A3-GCN, which incorporates a broader set of agreed nodes for training, has a lower pseudo-label accuracy (90\%) but higher overall accuracy (85.40\%), suggesting that it is better at generalizing across a more diverse set of nodes, potentially mitigating the effects of a distribution shift. The findings highlight how restricting the training set to high-confidence nodes may inadvertently exacerbate distribution shift, making the model more sensitive to certain node configurations at the cost of generalization.
\subsection{Individual models vs consensus model}
\begin{figure}[!ht]
	\begin{minipage}[b]{0.45\linewidth}
		\centering
		\centerline{\includegraphics[width=8cm]{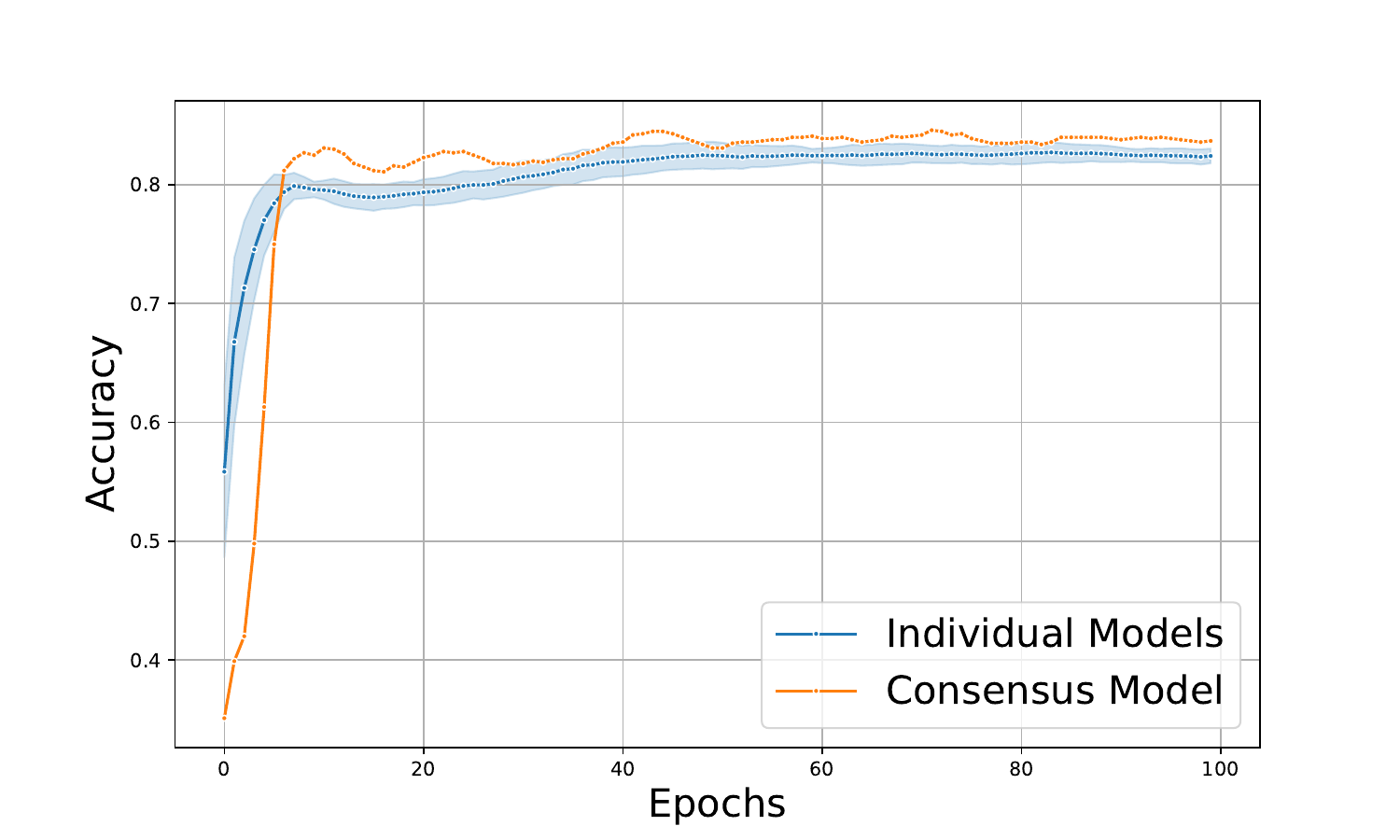}}
		\centerline{(a) Cora}\medskip
	\end{minipage}
	\hfill
	\begin{minipage}[b]{0.45\linewidth}
		\centering
		\centerline{\includegraphics[width=8cm]{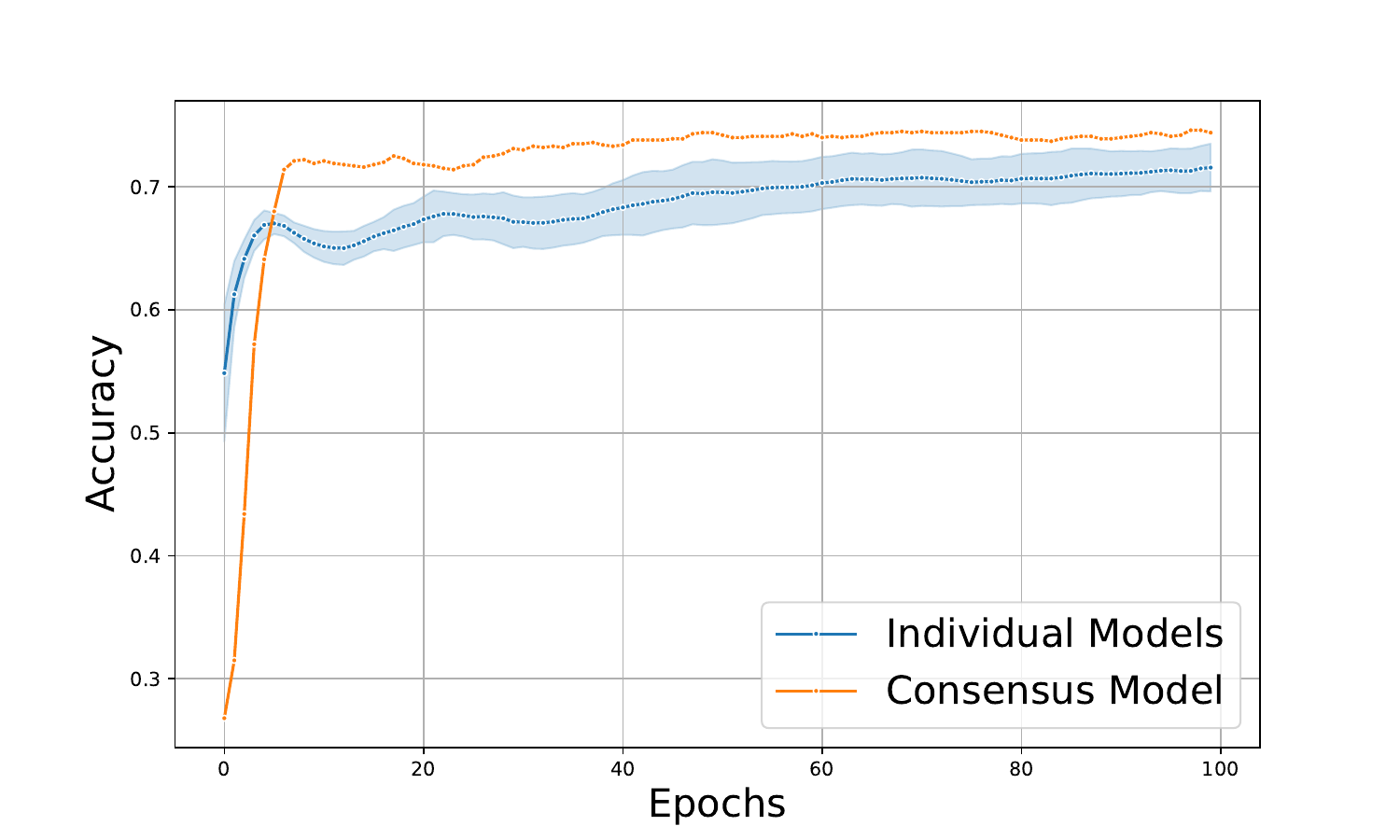}}
		\centerline{(b) Citeseer}\medskip
	\end{minipage}
	\hfill
	\begin{minipage}[b]{1\linewidth}
		\centering
		\centerline{\includegraphics[width=8cm]{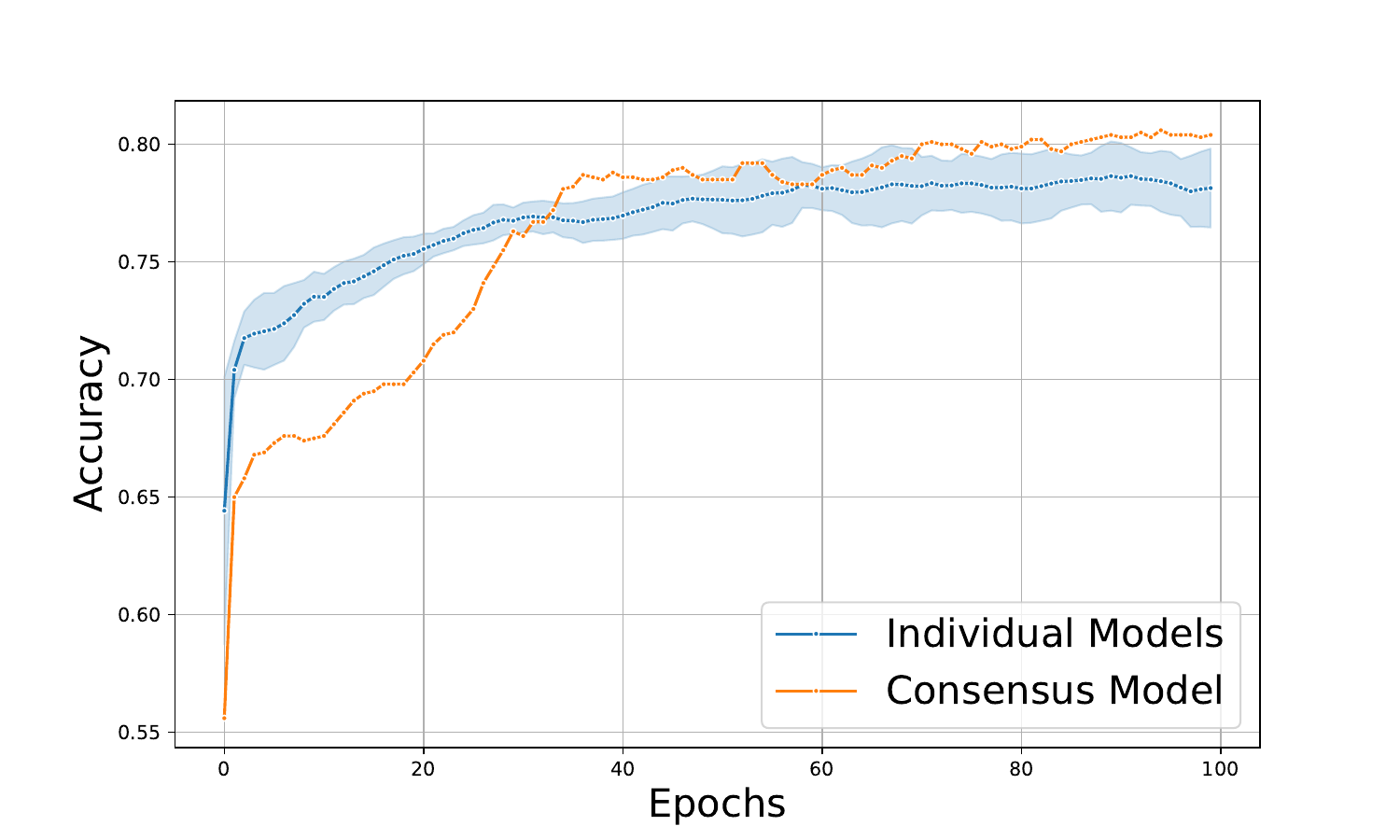}}
		\centerline{(c) PubMed}\medskip
	\end{minipage}
	\caption{Comparison of the accuracy of individual models vs the consensus model in each epoch for (a) Cora, (b) Citeseer and (c) Pubmed dataset.}
	\label{individual_vs_consensus}
\end{figure}

This experiment compares the performance of individual GCN models with a consensus model on three dataset. The average accuracy of the 10 individual models was calculated and plotted, along with error bands (standard deviation) representing the variance across models. The result is shown in Figure~\ref{individual_vs_consensus}.
We observe that the consensus model provides a more stable and reliable prediction compared to any of the individual models, demonstrating the benefit of aggregating multiple models' predictions.

\label{adaptive_vs_regular}
\subsection{Comparison with state-of-the-art}
We have compared our proposed algorithm with several state-of-the-art approaches in the SSL and self-training literature, particularly for graph-structured data in Table~\ref{tab:comparison}. To ensure the robustness and reliability of our results, we report performance averaged over 10 independent trials.

We observe that:
\begin{itemize}
	\item A3-GCN delivers competitive performance across all three datasets.
	\item IFC-GCN~\cite{hu2021rectifying} has the highest average accuracy on the PubMed dataset but exhibits a notably high standard deviation.
	\item GraphMix~\cite{verma2021graphmix} performs slightly better on Citeseer and PubMed (with higher variance) but significantly underperforms on the Cora dataset.
\end{itemize}
\small
\begin{table}
	\centering
    
	\caption{Comparison of SSL-based approaches for node classification. The highest accuracy is highlighted in bold.}
	\label{tab:comparison}
		\begin{tabular}{|l|c|c|c|c|c|c|c|c|}
			\hline
			& \textbf{GCN\cite{kipf2016semi}} & \textbf{SGC\cite{wu2019simplifying}} & \textbf{P-reg}\cite{yang2021rethinking} & \textbf{IFC-GCN}\cite{hu2021rectifying} & \textbf{GAM}\cite{stretcu2019graph} & \textbf{GraphMix}\cite{verma2021graphmix} & \textbf{Cautious}\cite{wang2023deep} & \textbf{A3-GCN}\\ \hline
			Cora \cite{mccallum2000automating} & 81.5 & 81 & 83.38  &  84.5$\pm$0.4 & 82.28$\pm$0.48 & 83.94$\pm$0.57 & 83.94$\pm$0.42  & \textbf{85.37$\pm$0.25}
			\\ \hline
			CiteSeer \cite{10.1145/276675.276685} &70.3 & 71.9 & \textbf{74.83} & 74.2$\pm$1.21 & 72.74$\pm$0.62 & 74.72$\pm$0.59 & 72.96 $\pm$ 0.22 & 74.36$\pm$0.31
			\\ \hline
			PubMed \cite{Sen_Namata_Bilgic_Getoor_Galligher_Eliassi-Rad_2008} & 79.0 & 78.9 & 80.11 & \textbf{81.3$\pm$0.93} & 79.60$\pm$0.63 & 80.98$\pm$0.55 & 79.98$\pm$0.92 &80.59$\pm$0.29\\ \hline                                
		\end{tabular}
        
\end{table}

\subsection{Ablation Study}
We investigate the effectiveness of adaptive strategies to improve the performance of the consensus-based model. Two key adaptive techniques are examined: (i) Adaptive confidence thresholding and (ii) Adaptive high confidence sampling. 
We conduct experiments on the Cora dataset using A3-GCN consensus setup with 10 individual models. The following configurations are evaluated:

\begin{itemize}
	\item \textbf{No Adaptive:} A predetermined confidence threshold $\theta_{\text{conf}}$ (either 0.99 or 0.95) is used, without any adaptive sampling criterion.  
	
	\item \textbf{Adaptive Confidence Thresholding:} The confidence threshold is dynamically updated at each epoch according to \eqref{adaptive_conf}, but no adaptive sampling is applied to high-confidence nodes.  
	
	\item \textbf{Adaptive Sampling Selection:} High-confidence data for each model is sampled based on the random adaptive criterion in \eqref{high-ratio}, while maintaining a fixed threshold.  
	
	\item \textbf{Combined Adaptive Techniques:} Both adaptive confidence thresholding and adaptive sampling selection are applied together.  
	
	\item \textbf{No Ensemble or Adaptive Learning:} No ensemble learning ($k=1$) is used, and no parameters undergo adaptive learning.  
	
\end{itemize}

The results across 10 trials are presented in Table \ref{tab:ablation}. We observe the following:

\begin{itemize} \item The results for \texttt{No Ensemble or Adaptive Learning} highlight the critical role of the proposed components in improving performance. The results in this setting are significantly lower compared to all other configurations. \item The results for \texttt{No Adaptive} with two thresholds of $\theta_{\text{conf}} = 0.99$ and $\theta_{\text{conf}} = 0.95$ demonstrate the sensitivity of the algorithm to the threshold value. This further emphasizes the importance of dynamic thresholding when selecting high-confidence pseudo-labels. \item A comparison between \texttt{Adaptive Confidence Thresholding} and \texttt{Combined Adaptive Techniques} suggests that adaptive threshold selection plays a more significant role in performance than adaptive sampling selection across the three datasets. However, adaptive sampling also provides some improvement, particularly for the Citeseer dataset. \end{itemize}
\begin{table}[h]
	\centering
	\caption{Ablation Study of the A3-GCN Model.}
		\begin{tabular}{|l|c|c|c|}
			\hline
			& \textbf{Cora} & \textbf{Citeseer} & \textbf{PubMed} \\ \hline
			No Adaptive ($\theta_{conf}=0.99$) & 84.36$\pm$0.49 & 73.41$\pm$0.86  & 77.59$\pm$0.46\\ \hline
			No Adaptive ($\theta_{conf}=0.95$) & 85.47$\pm$0.63 & 73.02$\pm$1.15&  79.35$\pm$0.43\\ \hline
			Adaptive Confidence Thresholding  & \textbf{85.47$\pm$0.30}  & 73.84$\pm$0.45 & 80.29$\pm$0.49 \\ \hline
			Adaptive Sampling Selection ($\theta_{conf}=0.99$) & 84.02$\pm$0.34 & 74.00$\pm$0.46  & 80.04$\pm$0.26 \\ \hline
			No Ensemble or adaptive learning & 82.98$\pm$1.03  & 71.67$\pm$1.86 & 75.82$\pm$2.09 \\ \hline
			Combined Adaptive Techniques (A3-GCN) & 85.37$\pm$0.25 & \textbf{74.36$\pm$0.31} & \textbf{80.59$\pm$0.29}\\ \hline               
		\end{tabular}
	\label{tab:ablation}
\end{table}

\subsection{Confidence threshold} 
\begin{figure}[!ht]
	\begin{minipage}[b]{0.45\linewidth}
		\centering
		\centerline{\includegraphics[width=8cm]{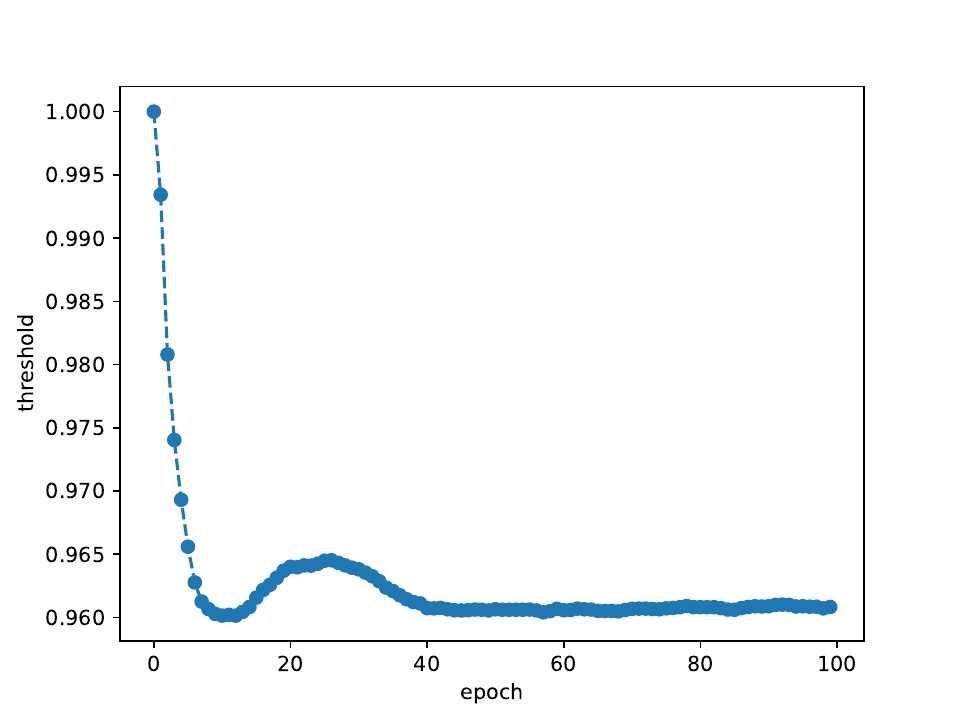}}
		\centerline{(a) Cora}\medskip
	\end{minipage}
	\hfill
	\begin{minipage}[b]{0.45\linewidth}
		\centering
		\centerline{\includegraphics[width=8cm]{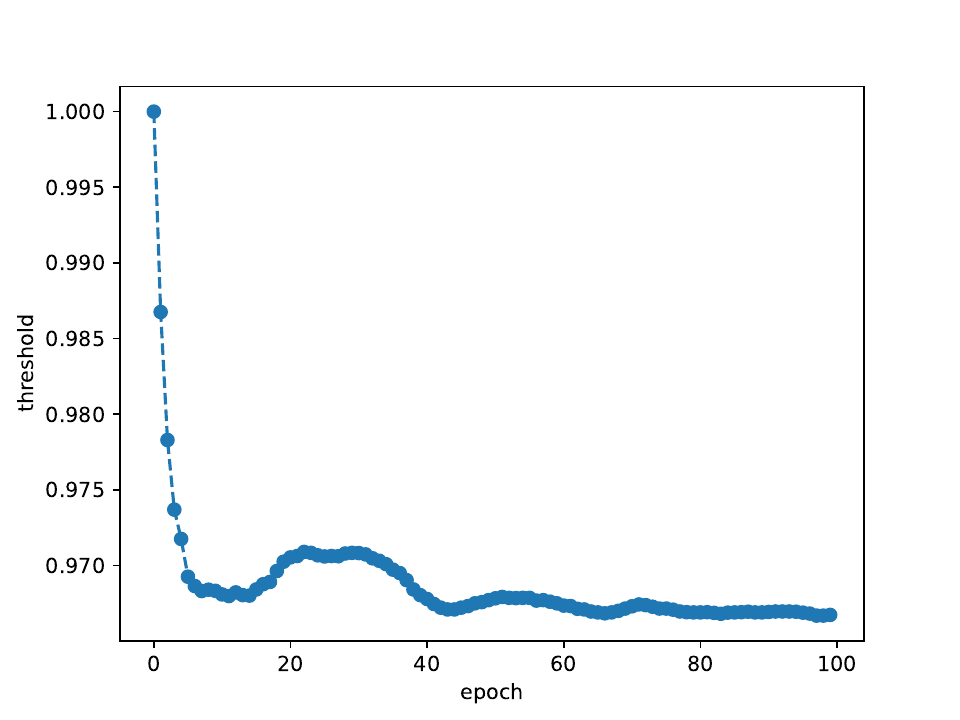}}
		\centerline{(b) Citeseer}\medskip
	\end{minipage}
	\hfill
	\begin{minipage}[b]{1\linewidth}
		\centering
		\centerline{\includegraphics[width=8cm]{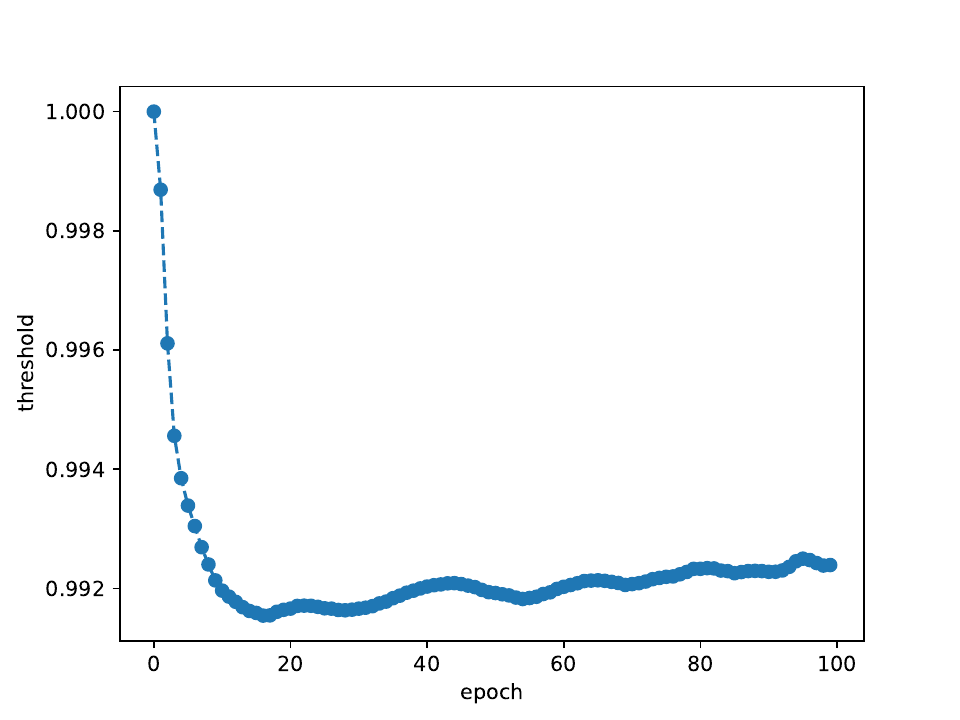}}
		\centerline{(c) PubMed}\medskip
	\end{minipage}
	\caption{The adaptive values of $\theta_{conf}$ across different epochs for various datasets typically decrease rapidly in the initial epochs and then converge to a stable value in the later epochs.}
	\label{threshold_adaptive}
\end{figure}

In this experiment, we present the evolution of the threshold $\theta_{\mathrm{conf}}$ across different epochs for three datasets.
The values of threshold $\theta_{\mathrm{conf}}$ for the Cora, Citeseer, and PubMed datasets are plotted in Figure~\ref{threshold_adaptive}.
We observe that the threshold rapidly decreases during the initial epochs, coinciding with a steady increase in agreement between the models, which then stabilizes in later epochs. This aligns with the `early stopping'\cite{bai2021understanding} concept in neural networks, where models first learn the clean patterns before overfitting to noisy ones. Consequently, we believe that high-confidence nodes in the early epochs are more reliable than those in later epochs, where the threshold should be reduced more cautiously. Additionally, GCN models tend to train quickly and achieve near-final accuracy early on, meaning that reliable high-confidence nodes are already available in the initial epochs.

\subsection{Sensitivity to the parameters}
\begin{figure}[!ht]
	\begin{minipage}[b]{0.45\linewidth}
		\centering
		\centerline{\includegraphics[width=8cm]{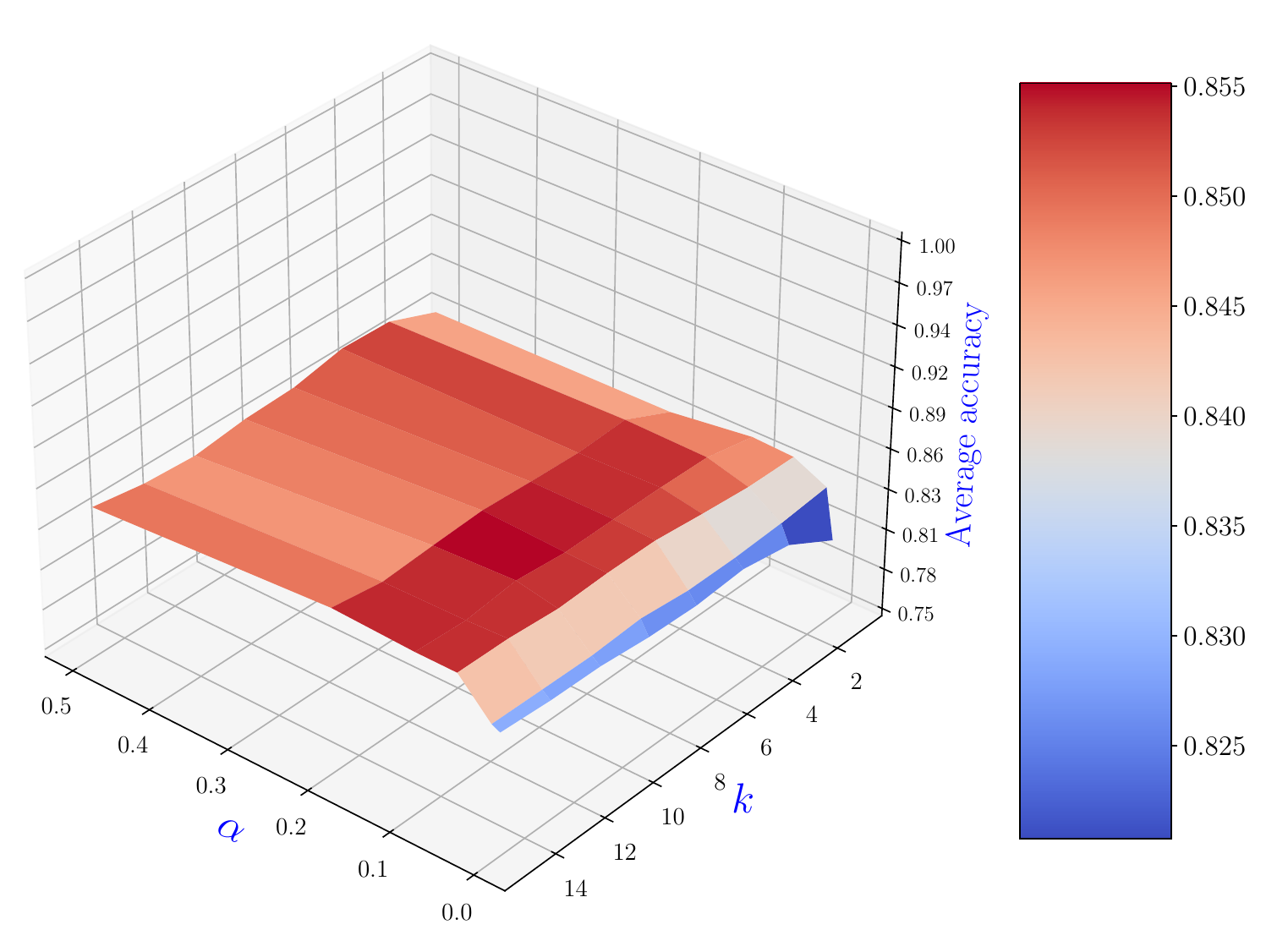}}
		\centerline{(a) $\alpha$ vs $k$}\medskip
	\end{minipage}
	\hfill
	\begin{minipage}[b]{0.45\linewidth}
		\centering
		\centerline{\includegraphics[width=8cm]{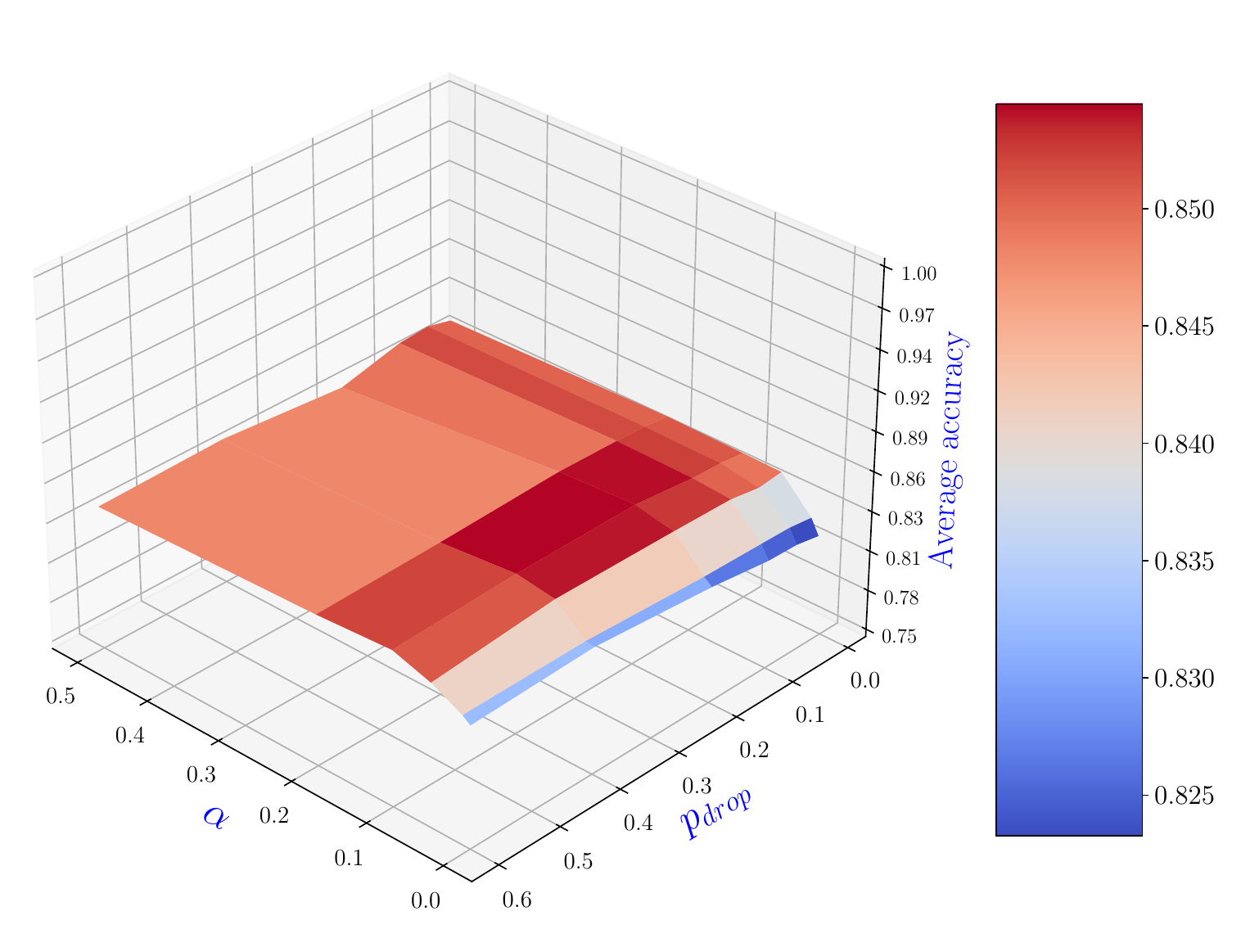}}
		\centerline{(b) $p_{drop}$ vs $\alpha$}\medskip
	\end{minipage}
	\hfill
	\begin{minipage}[b]{1\linewidth}
		\centering
		\centerline{\includegraphics[width=8cm]{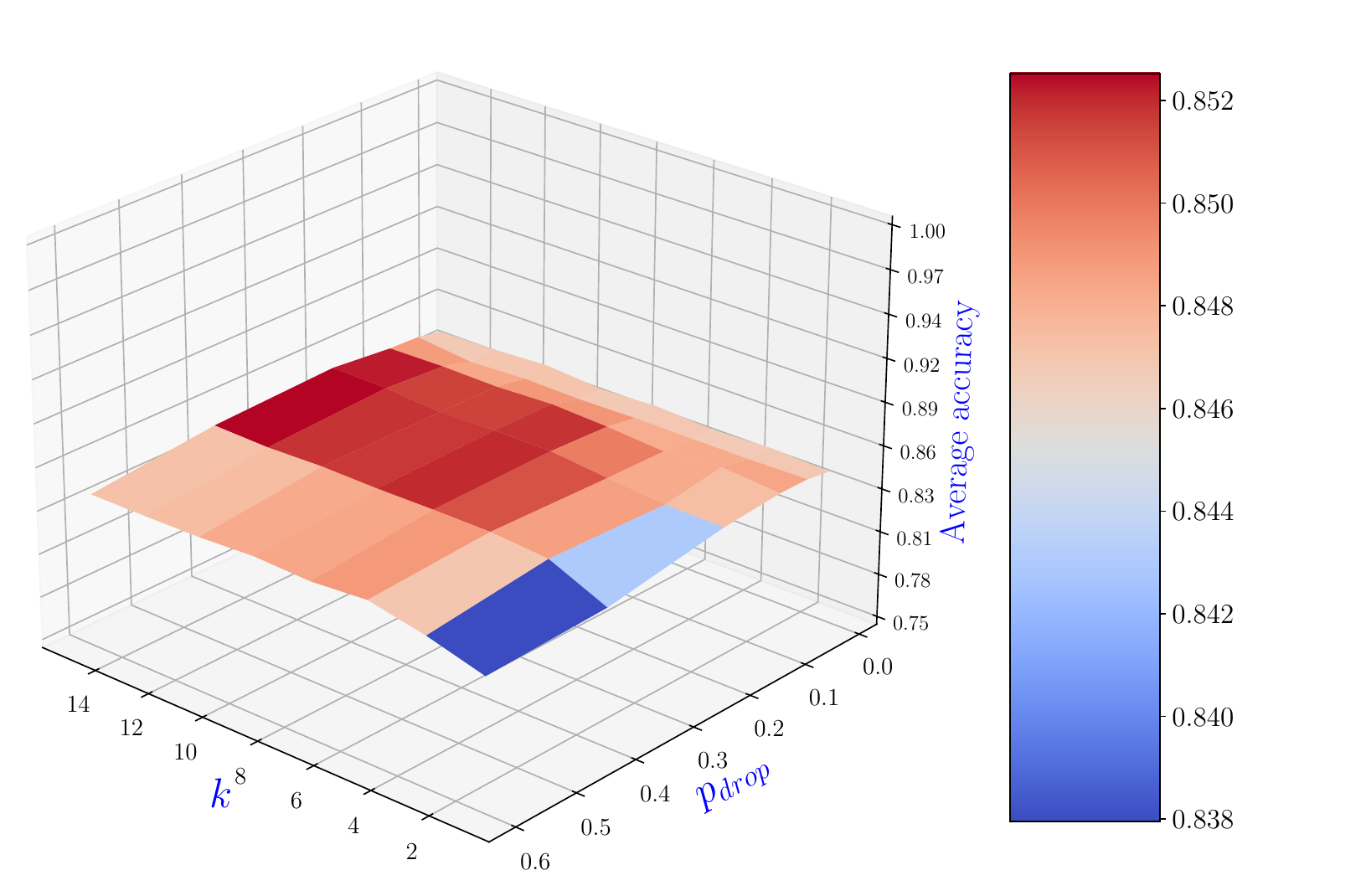}}
		\centerline{(c) $p_{drop}$ vs $k$}\medskip
	\end{minipage}
	
	\caption{Sensitivity to different parameters. Average accuracy across (a) varying values of \( \alpha \) (learning rate) and \( k \) (number of ensemble models), (b) \( p_{\text{drop}} \) (probability of dropping edges in graph augmentation) and \( \alpha \), and (c) \( p_{\text{drop}} \) and \( k \).}
	\label{k_alpha}
\end{figure}

We evaluate the performance of A3-GCN for various parameter settings. Specifically, we explored the following parameter configurations: the number of multi-models \( k \in \{1, 3, 5, 7, 9, 11, 13, 15\} \), the edge drop probability used in generating augmented graphs \( p_{\text{drop}} \in \{0.01, 0.05, 0.1, 0.2, 0.4, 0.6\} \), and the learning rate for the adaptive confidence threshold \( \alpha \in \{0, 0.01, 0.05, 0.1, 0.2, 0.5\} \). The performance was evaluated across these different configurations, and the results are summarized in Figure~\ref{k_alpha}. Our observations are as follows:
\begin{itemize}
	\item The performance of A3-GCN remains stable across different values of \( k \), provided \( k \) is not too small (i.e., less than 5).
	\item The optimal range for \( \alpha \) lies within \( [0.05, 0.2] \). Smaller values of \( \alpha \) result in overly conservative confidence, retaining only the most confident nodes, while excessively large values cause a rapid decrease in the confidence threshold, leading to the selection of unreliable samples as pseudo-labels.
	\item The optimal range for \( p_{\text{drop}} \) is between \( [0.1, 0.4] \). This is expected, as very small values of \( p_{\text{drop}} \) produce similar graphs (models), while excessively large values lead to a significant loss of connectivity information.
	
\end{itemize}
\subsection{Performance under limited supervision}
\begin{figure}[h]
	\begin{minipage}[b]{1\linewidth}
		\centering
		\centerline{\includegraphics[width=14cm]{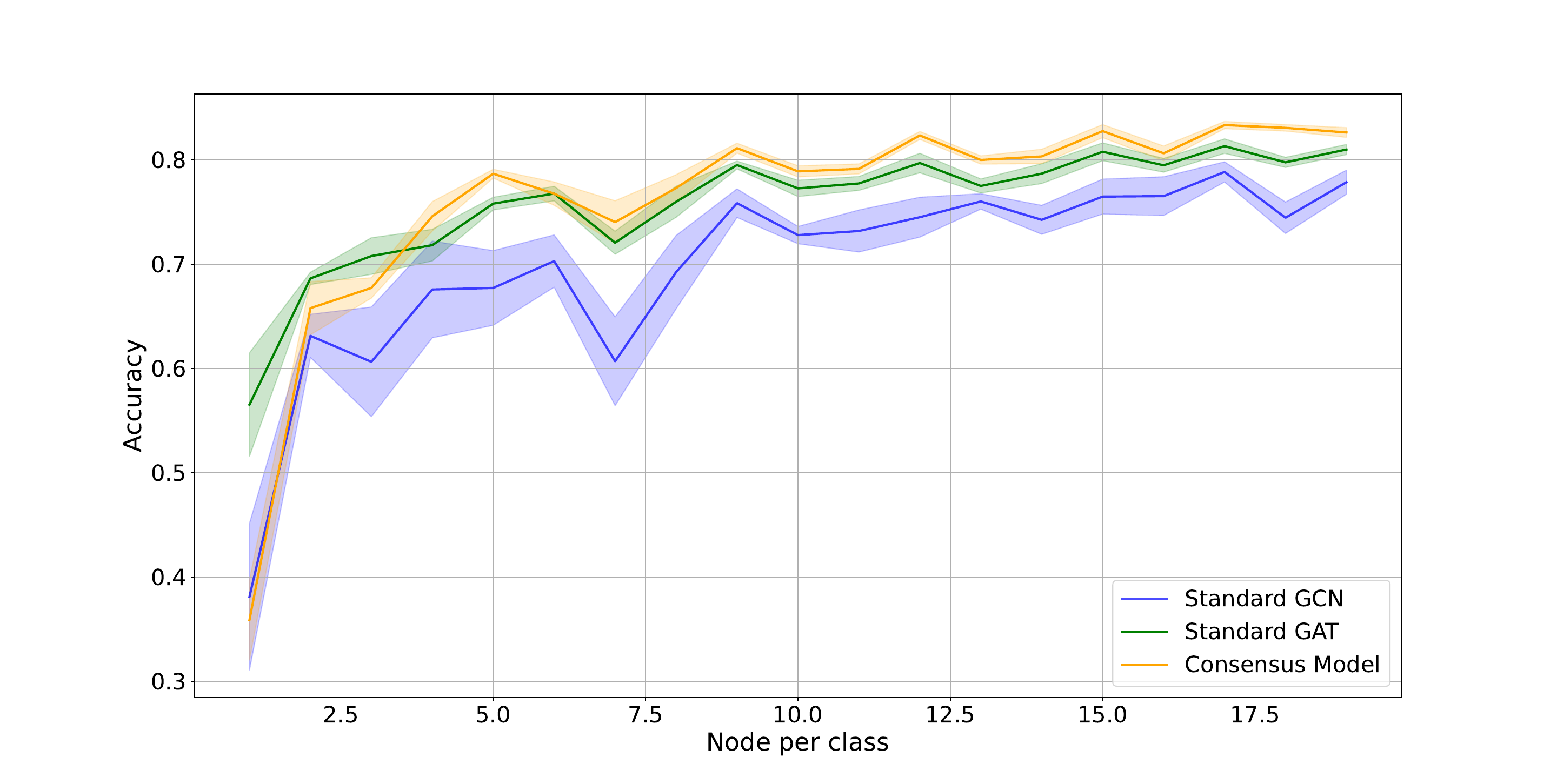}}
	\end{minipage}
	\caption{Performance with respect to different label rates (number of labeled nodes per class)}
	\label{label-rate}
\end{figure}

To analyze the model?s performance under different levels of supervision, we conducted an experiment using the Cora dataset. We randomly selected a fixed number of nodes per class and assigned their ground-truth labels, using them as the training set. The remaining nodes were used to assess the model?s generalization performance.
By varying the number of labeled nodes per class, we analyzed A3-GCN's performance in a limited supervision setting.
The average accuracy over 10 trials for different numbers of labeled nodes per class is shown in Figure~\ref{label-rate}. We compared the performance of A3-GCN against standard GCN and GAT. A3-GCN consistently outperforms standard GCN and, except in cases of extremely low supervision, also surpasses standard GAT. These results further validate the efficiency of the proposed ensemble approach in limited supervision settings.
\subsection{Robustness to noisy graph structures}
\begin{figure}[h]
	\begin{minipage}[b]{1\linewidth}
		\centering
		\centerline{\includegraphics[width=14cm]{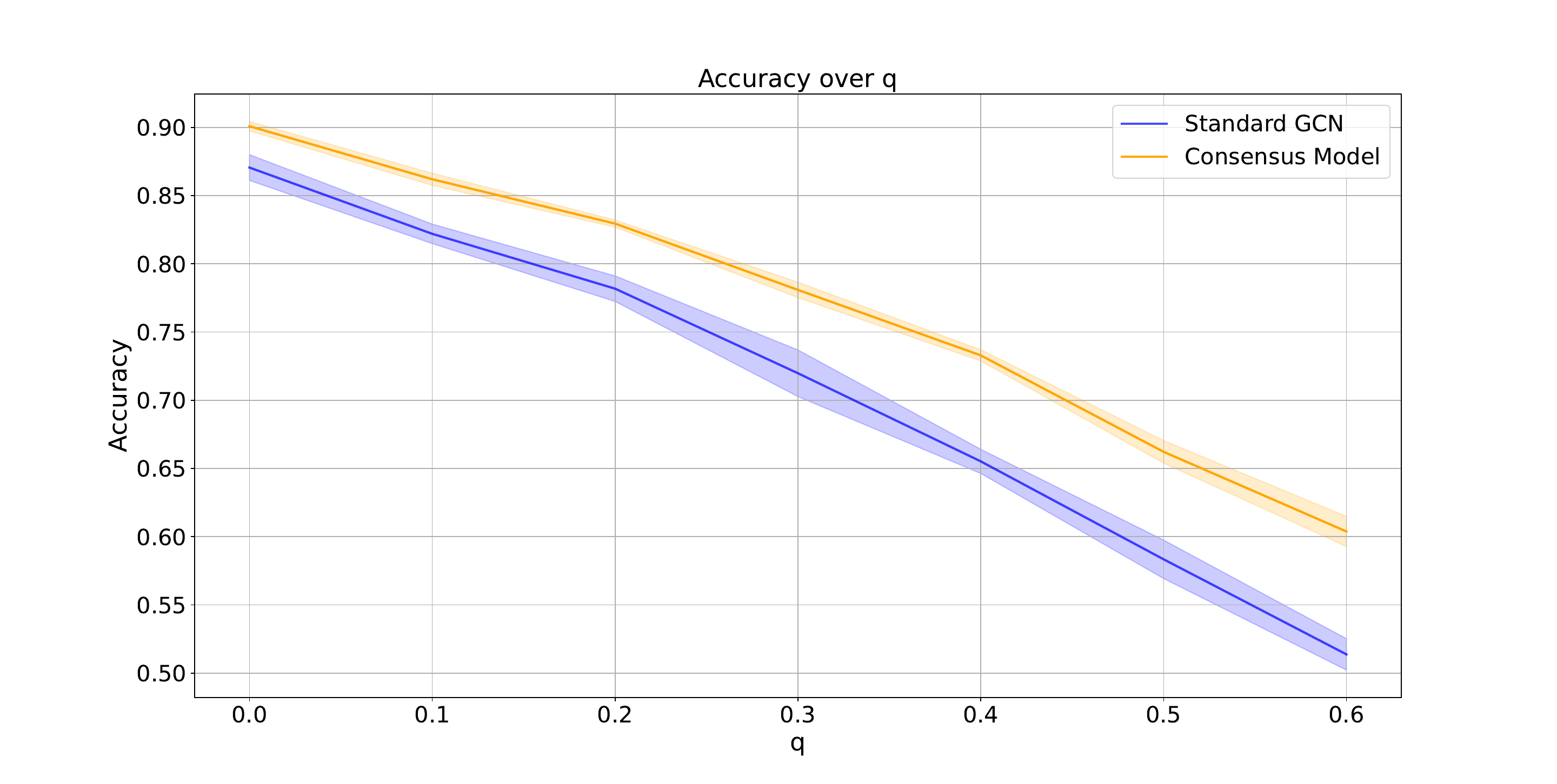}}
	\end{minipage}
	\caption{Performance with respect to increasing noisy graph structure}
	\label{noisy_graph}
\end{figure}

In this experiment, we used the Cora dataset, beginning with the removal of all noisy edges. Noisy edges are defined as those that connect nodes from different classes, which could introduce confusion during the learning process. At each step, we introduced an increasing proportion of noisy edges, represented by $q$ in the Figure~\ref{noisy_graph}, simulating a scenario where the graph structure becomes increasingly unreliable.

After each addition of noisy edges, we evaluated the accuracy of our model and compared it with the performance of a standard GCN under the same conditions. This setup allows us to assess how the model's accuracy is affected by the presence of noisy edges and how well our model can maintain performance compared to the standard GCN when faced with increasingly noisy graph structures.
The average of 10 trials is shown in Figure~\ref{noisy_graph}. A3-GCN's superior performance over standard GCN is notably evident, highlighting the effectiveness of ensemble learning and graph augmentation in handling noisy graph structures.
\section{Conclusion}\label{sec6}


In this paper, we introduced a robust ensemble-learning approach for GCNs, utilizing the principles of adaptation, agreement, and aggregation to enhance semi-supervised node classification in noisy, real-world graph datasets. By generating diverse graph views through edge drop augmentation, our method mitigated the impact of noisy edges, enhancing the model's ability to adapt and learn from varied graph structures. The use of adaptive high-confidence pseudo-labels, dynamic thresholding and model agreement ensured that the pseudo-labeling process remained reliable, reducing the risk of confirmation bias while maintaining sufficient diversity. Furthermore, the consensus model, trained on nodes with strong agreement across multiple models, stabilized predictions and prevented overfitting to noisy pseudo-labels. These components worked synergistically to create a more robust, adaptable, and efficient graph model, particularly for semi-supervised learning tasks. Experimental results on real-world datasets validated the effectiveness of our approach, demonstrating its potential for advancing graph-based learning, especially in noisy data settings.





\bibliography{sn-bibliography}

\end{document}